\documentclass{article}
\usepackage{spconf,amsmath,epsfig}

\usepackage{amssymb}
\usepackage{booktabs, cite, color}
\usepackage{pifont} 
\usepackage{algorithmic, adjustbox}
\usepackage{algorithm}

\usepackage{graphicx}
\usepackage{subcaption}
\usepackage{cleveref}   

\let\OLDthebibliography\thebibliography
\renewcommand\thebibliography[1]{
  \OLDthebibliography{#1}
  \setlength{\parskip}{0pt}
  \setlength{\itemsep}{0pt plus 0.3ex}
}

\begin{document}\sloppy

\def\x{{\mathbf x}}
\def\L{{\cal L}}

\title{FedGCA: Global Consistent Augmentation Based Single-Source Federated Domain Generalization}


\name{Yuan Liu\textsuperscript{1}, Shu Wang\textsuperscript{1}, Zhe Qu\textsuperscript{1}, Xingyu Li\textsuperscript{2}, Shichao Kan\textsuperscript{1}, Jianxin Wang\textsuperscript{1*} 
}
\address{\textsuperscript{1}Hunan Provincial Key Lab on Bioinformatics, School of Computer Science and Engineering,\\ Central South University, Changsha, Hunan, China\\\textsuperscript{2}Tulane University, New Orleans, USA}




\maketitle

\begin{abstract}
Federated Domain Generalization (FedDG) aims to train the global model for generalization ability to unseen domains with multi-domain training samples. However, clients in federated learning networks are often confined to a single, non-IID domain due to inherent sampling and temporal limitations. The lack of cross-domain interaction and the in-domain divergence impede the learning of domain-common features and limit the effectiveness of existing FedDG, referred to as the single-source FedDG (sFedDG) problem. To address this, we introduce the Federated Global Consistent Augmentation (FedGCA) method, which incorporates a style-complement module to augment data samples with diverse domain styles. To ensure the effective integration of augmented samples, FedGCA employs both global guided semantic consistency and class consistency, mitigating inconsistencies from local semantics within individual clients and classes across multiple clients. The conducted extensive experiments demonstrate the superiority of FedGCA.

\end{abstract}
\section{Introduction}
\label{sec:intro}
In recent years, Federated Learning (FL) has emerged as a potent paradigm for effectively training machine learning models across a network of decentralized clients \cite{mcmahan2017communication, li2021model, acar2021federated, qu2022generalized}. Numerous FL applications are deployed in scenarios where data decentralization is imperative due to privacy concerns, communication limitations, or regulatory constraints. In the earlier stages of FL research \cite{mcmahan2017communication, li2021model, acar2021federated, qu2022generalized}, there was a common assumption that the training and testing data of clients followed the same distribution, ensuring the generalizability of the global model to test data. However, guaranteeing that a trained model will always be applied to the distribution it was trained on is challenging. This inherent uncertainty necessitates a critical extension of FL known as Federated Domain Generalization (FedDG) \cite{liu2021feddg, zhang2021federated, nguyen2022fedsr}, which seeks to empower the global model to generalize to out-of-distribution (OOD) \cite{shen2021towards, qu2023prevent} domains, where the boundaries between different domains are unavailable, making the task challenging to detect and address.

\begin{figure}[t!]
  \centering
  \includegraphics[width=0.95\columnwidth]{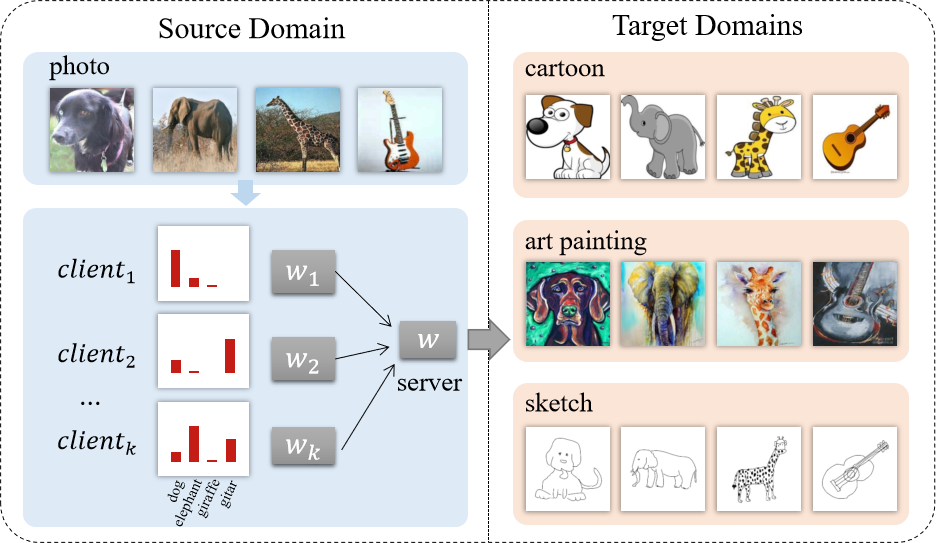}
  \caption{Description of the sFedDG problem.}
  \label{Fig:sFedDGProblem}
\end{figure}

Many existing FedDG studies \cite{liu2021feddg, zhang2021federated, nguyen2022fedsr} predominantly focus on scenarios where the clients in the FL network originate from multiple domains. These studies assume a one-to-one correspondence between clients and domains, leading to cross-domain concept drift among clients. Typically, they benefit from the rich domain information, facilitating straightforward generalization to unseen target domains. However, a more realistic scenario is where client samples are confined to specific times or regions. For instance, in seasonal flu prediction, samples exhibit distinct temporal feature distributions. Similarly, in image classification tasks within the desert environment, the images are prone to exhibit common background features associated with desert characteristics.

In such cases, clients are confined to a single domain, generating a new problem called \textbf{single-source Federated Domain Generalization (sFedDG)}. As illustrated in Fig.~\ref{Fig:sFedDGProblem}, all clients belong to one domain, but they aim to train a global model that can work for other unseen target domains. This issue entails two prominent challenges: 1) limited accessible domain styles. Unlike FedDG setups \cite{liu2021feddg, zhang2021federated, nguyen2022fedsr, qu2023convergence}, where clients engage with multiple domains, the constraint of sFedDG restricts clients' ability to interact with and learn from other domains. This limitation hinders the exploration of common features to achieve effective domain generalization. Furthermore, sFedDG can be considered as an advanced extension of the OOD testing problem, focusing on generality enhancement \cite{shen2021towards}. 2) Semantic inconsistency on both local and global objectives. Unlike the single-source Domain Generalization (sDG) problem \cite{wang2021learning, cugu2022attention, peng2022out}, the data distribution on each client exhibits non-IID nature in sFedDG due to the nature of FL. Despite the application of data augmentation techniques \cite{xu2020robust, zhao2020maximum} to diversify domain styles, it remains challenging due to the absence of samples for certain classes, resulting in an inconsistent decision boundary.

To address the introduced challenges, we present a novel and versatile algorithm: {\bf Federated Global Consistent Augmentation (FedGCA)}, crafted to elevate generalization performance against sFedDG. First, FedGCA employs a style-complement module, leveraging cutting-edge data augmentation techniques to effectively enrich data samples across a limited set of accessible domain styles, drawing inspiration from \cite{xu2020robust, zhao2020maximum, cugu2022attention}. Subsequently, FedGCA addresses the in-domain divergence caused by the non-IID nature of FL and ensures semantic information consistency across individual clients. It achieves this by incorporating both global guided semantic consistency and class consistency loss into the client training process. These losses utilize dynamic regularization strategies within the federated framework, effectively reducing features that are semantically irrelevant to the specific class distributions of clients.


The contributions of this paper are: 1) To the best of our knowledge, this is the first study to introduce and address the practical problem sFedDG, which poses two significant challenges that existing studies fail to address. 2) Innovatively addressing the unique challenges of sFedDG, we propose the corresponding FedGCA method: The style-complement module generates diverse and informative samples from the source domain with limited styles, and the two innovative strategies mitigate inconsistencies, thereby enhancing the model's ability to generalize effectively from one single domain to various domains. 3) Experimental results unequivocally show that FedGCA outperforms benchmarks on several datasets, underscoring its superior effectiveness.

\section{Preliminaries and Methodology}
\subsection{Preliminaries}
In FL, the primary objective is to collaboratively train a consensus global model. Typically, consider $K$ clients, denoted as $i \in [K]$, with non-IID datasets. Let $\mathcal{D}_i = \{(x_j^{(i)}, y_j^{(i)})\}_{j=1}^{N_i}$ be the dataset on client $i$, where $x_j^{(i)}$ is the $j$-th input sample and $y_j^{(i)}$ is its corresponding label. The size of datasets on client $i$ is $N_i$, and the total number of data samples across all clients is $N = \sum_{i=1}^{K}N_i$. Let $w$ be the global model, and $\mathcal{L}_i (w)$ be the local empirical risk function on client $i$. The objective of FL can be formulated as $\min_w \mathcal{L}(w) = \sum_{i=1}^{K} \frac{N_i}{N} \mathcal{L}_i(w)$. Each client calculates $\mathcal{L}_i (w)$ locally and then sends the updated model parameter $w_i$ to the server. The server then performs aggregation to update the global model $w$. The data distribution $p_i(x, y)$ of client $i$ are requested to be relevant (sampled from a family $\mathcal{E}$ of distributions).

In FedDG, the data distribution $p_i(x,y), \forall i$ is set to be different source domains, aiming to minimize the loss on unseen target domains $p_{\text{test}}(x,y) \sim \mathcal{E}$. The average case of this loss can be defined as $\mathbb{E}_{p_{\text{test}} \sim \mathcal{E}}[\mathbb{E}_{p_{\text{test}} (x,y)} \ell(w; x,y)]
$, where $\ell (w ;x, y)$ denotes the loss function of a data point $(x, y)$.
Note that in the DG/FedDG problem, different domains are typically considered to have a significant shift in style, texture, or appearance \cite{nguyen2022fedsr, liu2021feddg}, while clients are set in the same label distribution across domains \cite{zhang2021federated}. 

\subsection{sFedDG}
In this paper, we consider a new problem called {\bf single—source Federated Domain Generalization (sFedDG)}. More specifically, given a source domain $\mathcal{S}$, data distributions $p_i(x, y)$ across the clients are sampled from the same domain $\mathcal{S}$, i.e., $p_i(x, y) \sim \mathcal{S}, \forall i \in [K]$ with $p_{i}(x,y) \neq p_{i'}(x,y) ,\forall i ,i' \in [K]$. The goal of sFedDG is to collaboratively learn a global model that can generalize to unseen target domains $\mathcal{T}$. The source domain $\mathcal{S}$ and target domains $\mathcal{T}$ belong to the family $\mathcal{E}$ of distributions. For the test dataset distribution $p_{\text{test}}(x, y) \sim \mathcal{T}$, $p_{\text{test}}(x, y) \ne p_i(x, y)$ and $\mathcal{T}$ cannot be accessed in training. The expectation on target domains can be defined as $\mathbb{E}_{p_{\text{test}} \sim \mathcal{T}}[\mathbb{E}_{p_{\text{test}} (x,y)}\ell(w; x,y)]$.

With no prior knowledge of the target domains $\mathcal{T}$, we can just estimate the objective using finite clients and finite samples from the source domain $\mathcal{S}$. As a result, the sFedDG objective function can be accessed as $\min_{w}\mathcal{L}^{DG} (w) \triangleq \min_{w} \mathbb{E}_{p_i\sim \mathcal{S}}[\mathbb{E}_{p_i (x,y)}\ell(w; x,y)]$.

\subsection{FedGCA}
\begin{figure*}[t]
  \centering
    \includegraphics[width=1.8\columnwidth]{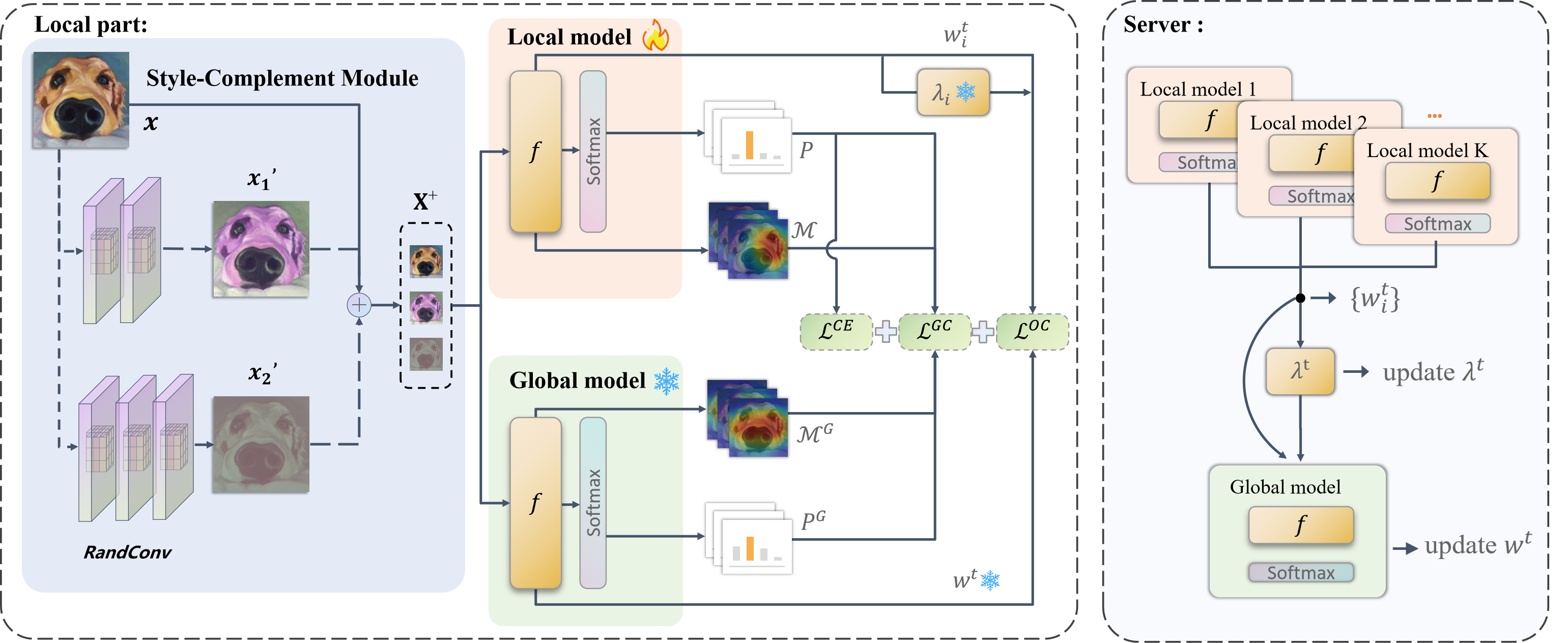}
    \caption{Overview of the FedGCA framework. \textbf{Left:} Local training for the client. \textbf{Right:} Global update on the server.}
    \label{fig:framwork_local}
  \hfill
\end{figure*}

The sFedDG problem primarily faces two challenges: (1) limited accessible source domain styles and (2) semantic inconsistency on both local and global objectives. To address these challenges, we propose a method called {\bf Federated Global Consistent Augmentation (FedGCA)}, illustrated in Fig.~\ref{fig:framwork_local}. Specifically, to tackle the first challenge, our proposed FedGCA method augments the source data samples through various transformations or generates pseudo-novel samples for the source domain. However, generated data introduces semantic noise, and clients can only generate enhanced data from their own distributions, leading to drift between clients and incurring the second challenge. Therefore, we first design semantic consistency loss from both coarse-grained and fine-grained perspectives, incorporating global information for constraint. Additionally, we employ dynamic regularization to guide augmented clients' training towards consistent objectives.

\textbf{Style-complement module.}
The style-complement module functions as either a data augmentation filter or a generation model \cite{xu2020robust, zhao2020maximum, cugu2022attention, sheng2023modeling}, aiming to enhance the data samples $x$ from the source domain $\mathcal{S}$ by synthesizing $x'$ with the same semantic information but different styles. Let $X^+=\{x ,x' \}$, with corresponding labels $Y^+=\{y ,y\}$. Following the acquisition of synthetic data samples, the objective of sFedDG can be formulated as follows:
\begin{equation}\label{Eq:obj_sFedDG_au}
\begin{split}
    \min_{w}\mathcal{L}^{DG} (w) \triangleq \min_{w} \mathbb{E}_{p_i \sim \mathcal{S}}[\mathbb{E}_{p_i (X^+,Y^+)}\ell(w; X^+,Y^+)].
\end{split}
\end{equation}
The aim is to enhance the learning of domain-invariant features, ultimately extending and refining the boundaries between classes. 
To augment images while preserving semantic information, the module incorporates variants that combine Random Texture\cite{xu2020robust}, a Mixing Variant of the AugMix strategy \cite{hendrycks2020augmix}, and Multi-scale Texture Corruption \cite{cugu2022attention}.

To increase the diversity, we performed some random repetitions of the RandConv operation \cite{xu2020robust, choi2023progressive} with the same parameters for the input to augment $J$ samples. As such, the set $X^+_i$ for client $i$ comprises the original data $x$ and the $J$ distinct augmented samples $X_i^+ = \{x_i, x'_{i,1},...,x'_{i,J}\}$.
However, the generated samples on client $i$ are based on the distribution $p_i (x,y)$, which lacks client semantic information from $p_{i'}(x,y)$ where $i \neq i' \in [K]$. Consequently, this introduces inconsistency in both semantics on a single client and the class distribution across all clients for the second challenge. Accordingly, we design the {\bf Global Guided Semantic Consistency} loss and the {\bf Class Consistency} loss. These losses aid the model in learning from $X^+_i$ with diverse information and consistency across clients, enriching it for unseen domains $\mathcal{T}$.

\textbf{Global Guided Semantic Consistency.} To ensure semantic consistency within client $i$, we employ a loss function $\mathcal{L}_i^{CP}$ that encourages consistent predictions for the same image but with different filters. Additionally, we leverage global predictions to guide consistent predictions, aiding different client models in learning semantically consistent features for the same class. Specifically, $P_i = \{p_{i,j}\}_{j=0}^{J+1}$ denotes the softmax predictions of the training model $w_i$ on $X^+_i$, and $P^G_i = \{p_{i,j}\}_{j=0}^{J+1}$ represents the global model predictions on $X^+_i$. In contrast to contrastive-based FL methods \cite{li2021model}, moving the current local model away from the old local model may be too drastic. Instead, we utilize the Kullback-Leibler (KL) divergence to minimize the global-local semantic consistency.
The calculation of $\mathcal{L}_i^{CP}$ is as follows:
\begin{equation}\label{Eq:loss_consistent_predictions}
     \mathcal{L}_i^{CP} =-\sum_{p_{i,j} \in P}p_{i,j}\log (\Bar{p_i})-(1-p_{i,j})\log(1-\Bar{p_i})
\end{equation}
where $\Bar{p_i}$ represents the average of $P_i \cup P^G_i$. The inclusion of $\mathcal{L}_i^{CP}$ in \eqref{Eq:loss_consistent_predictions} serves to alleviate potential semantic information distortion by constraining the distribution shift of samples belonging to the same class, taking into account both local and global model considerations.

\begin{algorithm}[t!]
    \caption{FedGCA.}
    \label{alg:FedGCA}
    \renewcommand{\algorithmicrequire}{\textbf{Input:}}
    \renewcommand{\algorithmicensure}{\textbf{Output:}}
    \begin{algorithmic}[1]
        \REQUIRE $w$, $w_i$, $T$, $K$, $\eta$, $\lambda_i$, $\lambda$, $\alpha$, and $\beta$ 
        \ENSURE global model $w^T$ 
        \STATE \textbf{Initialization} : $w = w^0, \lambda_i =0, \lambda = 0$
        \FOR{$t =0, \dots, T-1$}
            \FOR{ each client $i\in\{0,1,..,K-1\}$ \textbf{in parallel} }
                    \STATE Download the $w^t$ from server as $w^t_{i,0}$
                    \FOR{$l \in\{0,1,..,I-1\}$}
                        \STATE Calculate $X^+$ by style-complement module
                        \STATE Calculate $\mathcal{L}_i^{DG}$ by \eqref{Eq:FedGCA_total_loss}
                        \STATE $w^t_{i,l+1}= w^t_{i,l} - \eta \nabla \mathcal{L}_i^{DG}$
                    \ENDFOR
                    \STATE $\lambda_i = \lambda_i-\alpha(w_{i, I-1}^t - w^t)$
                    \STATE Upload $w_{i}^t=w_{i, I-1}^t$ to server
            \ENDFOR
        \STATE $\lambda^{t+1} = \lambda^{t} -\frac{\alpha}{K}\sum_{i \in [K]}(w_i^t-w^t)$
        \STATE $w^{t+1}=\frac{1}{K}\sum_{i \in [K]}w_i^t-\frac{1}{\alpha}\lambda^{t+1}$
        \ENDFOR 
    \end{algorithmic}
\end{algorithm}

As $\mathcal{L}_i^{CP}$ primarily addresses coarse consistency, a generalizable model should focus on the same image regions regardless of their particular style. To enhance specific regions containing the most crucial semantic information, i.e., fine-grained consistency, we turn to Class Activation Maps (CAMs) \cite{cugu2022attention, zhu2021residual}. CAMs provide visual explanations for a given model's predictions by visualizing spatial regions that significantly contributed to the output in a particular feature map. Let $F_l(x, y)$ represent the activation of the $l$-th feature map at spatial location $(x,y)$ in the final convolutional layer, and $W_l$ represent the weight corresponding to the $l$-th feature map in the fully connected layer for a specific class. The $\text{CAM}$ for that class is calculated as $M= \sum_{l} W_l \cdot F_l(x)$. Similar to $\mathcal{L}^{CP}_i$, we compute the CAMs for $X^+_i$ using both the current training model and the old global model in client $i$, denoted by: $\mathcal{M}_i$ and $\mathcal{M}_i^G$ respectively, and calculate the KL divergence between $\mathcal{M}_i$ and their averages $\Bar{M}_i$ of $\mathcal{M}_i \cup \mathcal{M}_i^G$:
\begin{equation}\label{Eq:cal_cam}
    \mathcal{L}_i^{CAM} = \sum_{{M}_{i,j} \in \mathcal{M}_i} KL({M}_{i,j}||\Bar{M}_i).
\end{equation}
$\mathcal{L}^{CP}$ and $\mathcal{L}^{CAM}$ concentrate on the overall and local information of images, respectively, showcasing distinct adaptabilities to the source domain style.
As highlighted in \cite{cugu2022attention}, $\mathcal{L}^{CP}$ focuses on the overall information consistency in simpler images, and $\mathcal{L}^{CAM}$ is more effective for information-rich images.
Consequently, we propose that the two losses should complement each other to enhance images with high quality, which leads to the definition of the global guided semantic consistency loss as follows:
\begin{equation}\label{Eq:lossCAM}
    \mathcal{L}_i^{GC} = \mathcal{L}_i^{CP} + \mathcal{L}_i^{CAM}.
\end{equation}

\textbf{Class Consistency.} The semantic consistency in \eqref{Eq:lossCAM} aims to learn similar features from existing classes within one client. However, due to the non-IID nature of FL, if a client lacks data samples from some specific classes, it cannot augment enough data samples to approach the objective in \eqref{Eq:obj_sFedDG_au} and may incur a blurred decision boundary. To mitigate the performance degradation caused by class imbalance, we employ a dynamic regularizer \cite{acar2021federated} on each client via a similar pattern as ADMM, to further improve the consistency from a global perspective. Specifically, we denote our class consistent regularizer on client $i$ as $\mathcal{L}^{OC}_i$, which is defined as follows:
\begin{equation}\label{Eq:L_OC}
    \mathcal{L}_i^{OC} = - \frac{1}{\alpha}\lambda_i^\top w_i + \frac{1}{2}{\left \| w_i - w^t \right \|}^2 ,
\end{equation}
where $\alpha$ is the hyper-parameter and $\lambda_i$ is a copy of the local gradient on client $i$. Intuitively, the first term in \eqref{Eq:L_OC} slows down the gradient changes during client model training, resulting in smoother gradient changes. The second term aligns the local model with the server model. This regularizer modifies the client loss to make the stationary points of the client risk consistent with the server model. 

In summary, the total local loss with cross-entropy loss  $\mathcal{L}_i^{CE}$ of client $i$ is defined as follows:  
\begin{equation}\label{Eq:FedGCA_total_loss}
\begin{split}
    \mathcal{L}^{DG}_i = \mathcal{L}_i^{CE} +\alpha\mathcal{L}_i^{OC} +\beta\mathcal{L}_i^{GC}.
\end{split}
\end{equation}
where $\beta$ is the hyper-parameter to balance the impact of $\mathcal{L}_i^{GC}$ in the loss function $\mathcal{L}^{DG}_i$.
Note that we update the client (gradient) state variable as $\lambda_i = \lambda_i-\alpha(w_{i, I}^t - w^t)$ after the number of $I$ local training epochs. At the end of $t+1$ round, we aggregate the local model $\{w_i\}_{i=0}^K$ to minimize the global model $w^{t+1}$, and update the server state variable $\lambda$. For a comprehensive understanding of the proposed FedGCA, Fig. 2 illustrates the training framework overview, and Algorithm 1 details the sequence of steps in the model update process.

\section{Experiment}

\subsection{Experimental Setups}
\noindent{\bf Datasets:} We conducted experiments on the following three datasets. (1) \textbf{Digits} consists of five distinct datasets: MNIST, SVHN, USPS, Synth, and MNIST-M \cite{nguyen2022fedsr}. (2) The \textbf{PACS} \cite{li2017deeper} dataset encompasses four domains: photo, art, cartoon, sketch. Each domain contains 224 × 224 images belonging to seven categories, posing challenges due to substantial style shifts among domains.

\noindent{\bf Compared methods:} The FL setting includes 10 clients on each dataset, where the non-IID is followed by Dirichlet distribution with parameter 0.3. To validate the efficiency of FedGCA, we compare the performance of FedGCA against several state-of-the-art FL and FedDG methods, including FedAVG \cite{mcmahan2017communication}, FedSAM \cite{qu2022generalized}, FedDyn \cite{acar2021federated}, Moon \cite{li2021model}, FedADG \cite{zhang2021federated}, and FedSR \cite{nguyen2022fedsr}. Additionally, we introduce augmented variants of each benchmark denoted by "+RC," which represents the inclusion of the RandConv variant. Due to the page limitation, we defer the detailed experimental setups and part of results in the supplementary materials.

\subsection{Performance Evaluation}

\begin{table}[t!]
    \centering
    \caption{sFedDG accuracy (\%) on digits dataset. Models are trained on {\it MNIST}.}
\adjustbox{width=1\columnwidth}{
\begin{tabular}{c|ccccc}
\hline
method & SVHN & USPS & Synth & MNIST-M & Avg. \\ \hline
FedAVG & 23.41 & 73.92 & 30.58 & 52.23 & 45.04 \\
FedADG & 24.55 & 73.39 & 34.94 & 52.96 & 46.46 \\
FedSR & 24.28 & 72.29 & 34.52 & 54.21 & 46.33 \\
FedSAM & 25.92 & 74.17 & 34.66 & 53.25 & 47.00 \\
FedDyn & 25.51 & 70.54 & 35.01 & 54.61 & 46.42 \\
Moon & 22.91 & 67.90 & 28.84 & 50.66 & 42.58 \\ \hline
FedAVG+RC & 29.77 & 71.56 & 43.66 & 66.05 & 52.76 \\
FedADG+RC & 37.09 & 76.02 & 50.42 & 75.58 & 59.78 \\
FedSR+RC & 35.40 & 79.46 & 46.26 & 75.29 & 59.10 \\
FedSAM+RC & 39.64 & 75.38 & 52.68 & 80.14 & 61.96 \\
FedDyn+RC & 38.30 & 68.92 & 52.75 & 83.94 & 60.98 \\
Moon+RC & 33.91 & 71.99 & 48.41 & 78.56 & 58.22 \\
FedGCA(ours) & \textbf{47.19} & \textbf{78.66} & \textbf{59.89} & \textbf{85.74} & \textbf{67.87} \\ \hline
\end{tabular}}
\label{Tab:acc_digit_dir0.3}
\end{table}

\begin{table}[t!]
    \centering
    \caption{sFedDG accuracy (\%) on PACS. Models are trained on {\it photo} and test on the rest of the domains.}
\begin{tabular}{c|cccc}
\hline
method & art & cartoon & sketch & Avg. \\ \hline
FedAVG & 43.26 & 18.65 & 25.18 & 29.36 \\
FedADG & 44.12 & 18.29 & 25.12 & 29.51 \\
FedSR & 51.81 & 17.62 & 26.93 & 32.12 \\
FedSAM & 54.24 & 20.42 & 25.73 & 33.13 \\
FedDyn & 44.75 & 16.94 & 26.16 & 29.95 \\
Moon & 51.11 & 14.76 & 21.89 & 29.59 \\ \hline
FedAVG+RC & 57.32 & 21.18 & 28.48 & 35.99 \\
FedADG+RC & 58.79 & 19.41 & 23.89 & 34.69 \\
FedSR+RC & 55.92 & 26.31 & 22.19 & 34.14 \\
FedSAM+RC & 52.47 & 28.01 & 30.21 & 36.56 \\
FedDyn+RC & 62.21 & 26.57 & 36.64 & 41.14 \\
Moon+RC & 60.09 & 28.83 & 29.47 & 39.13 \\
FedGCA(ours) & \textbf{65.01} & \textbf{30.91} & \textbf{41.19} & \textbf{45.70} \\ \hline
\end{tabular}
\label{Tab:AblationDigits_office}
\end{table}

{\bf Accuracy Evaluation.} The comparison results on Digits between FedGCA and other methods are presented in Table~\ref{Tab:acc_digit_dir0.3}. It is evident that the performance of existing methods significantly improves upon the addition of RandConv (RC), underscoring the significance of data augmentation in addressing the sFedDG problem. Notably, FedGCA demonstrates an average performance gain of at least 5.91\% compared to all other methods. Specifically, FedGCA outperforms the second-best method, FedSAM+RC, by 7.55\%, 3.28\%, 7.21\%, and 5.60\% on test domains, respectively, highlighting its superiority in overall generalizability. Similarly, the comparison results on PACS are presented in Table~\ref{Tab:AblationDigits_office}, which can improve up to 21.74\%. While FedADG and FedSR have demonstrated effectiveness in the FedDG problem \cite{zhang2021federated, nguyen2022fedsr}, they show performance degradation in sFedDG due to the absence of multiple source domain styles. Our method consistently achieves the highest performance.

\begin{figure}[t!]
  \centering
  \begin{subfigure}{0.48\columnwidth}
    \includegraphics[width=0.92\columnwidth, height=0.67\columnwidth]{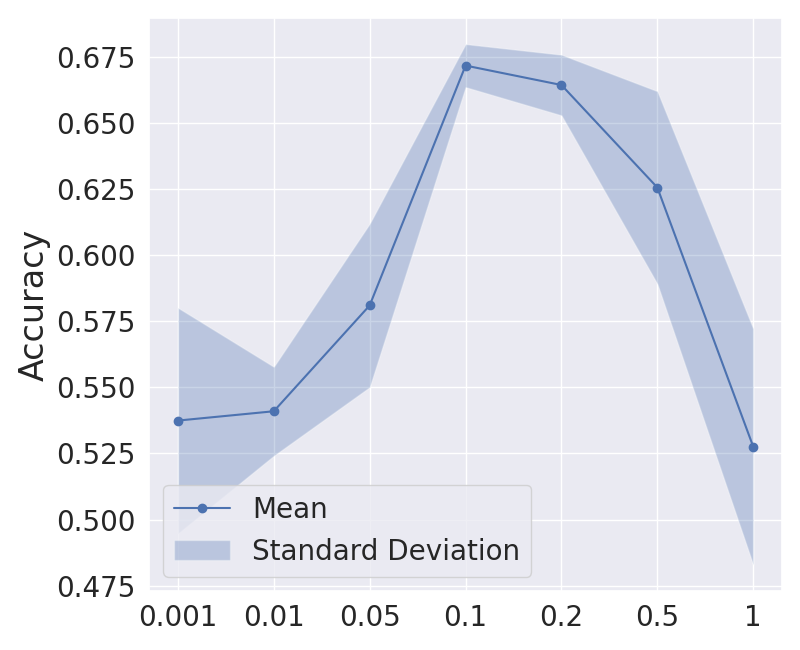}
    \caption{Sensitivity for $\alpha$ on Digits}
    \label{fig:short-alpha-d}
  \end{subfigure}
    \hfill
  \begin{subfigure}{0.48\columnwidth}
    \includegraphics[width=0.92\columnwidth, height=0.67\columnwidth]{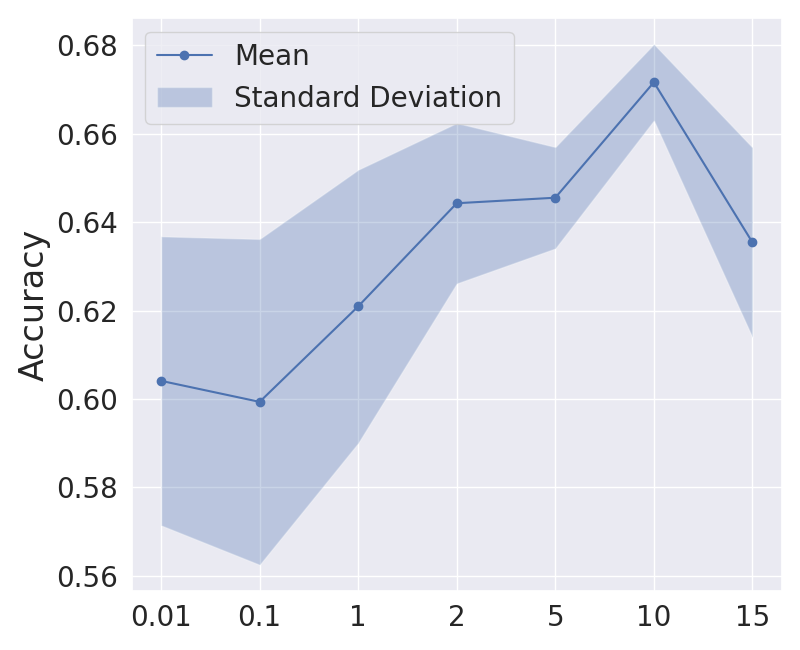}
    \caption{Sensitivity for $\beta$ on Digits.}
    \label{fig:short-beta-d}
  \end{subfigure}
  \begin{subfigure}{0.48\columnwidth}
    \includegraphics[width=0.92\columnwidth, height=0.67\columnwidth]{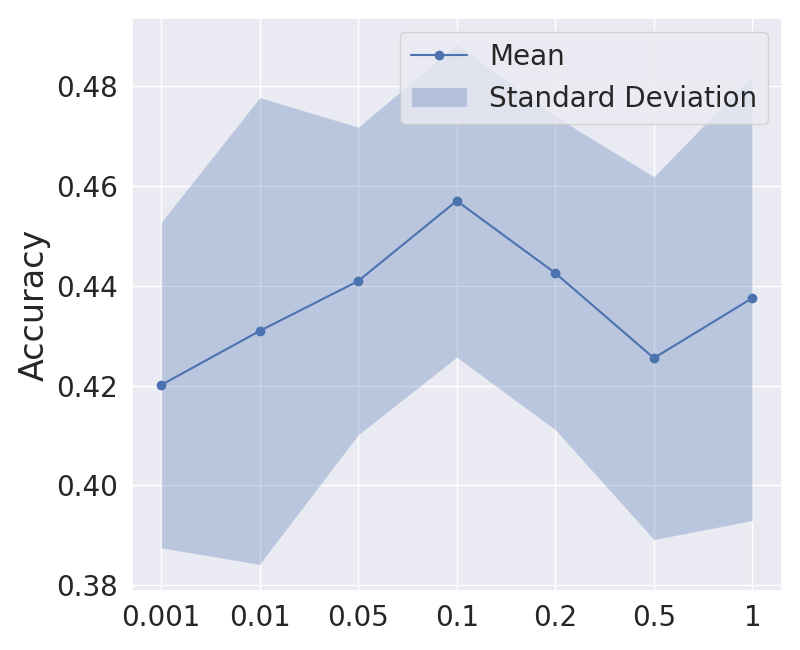}
    \caption{Sensitivity for $\alpha$ on PACS.}
    \label{fig:short-alpha-p}
  \end{subfigure}
   \hfill
  \begin{subfigure}{0.48\columnwidth}
    \includegraphics[width=0.92\columnwidth, height=0.67\columnwidth]{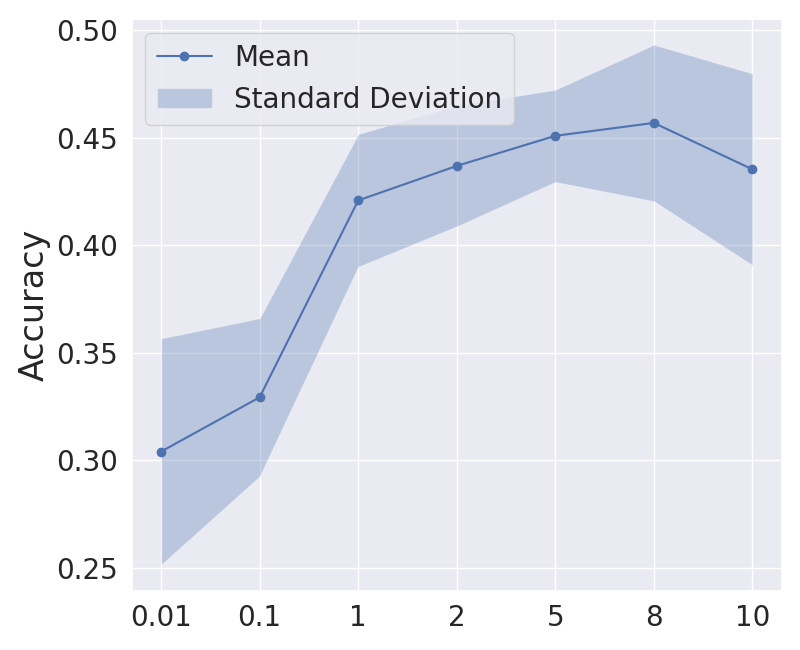}
    \caption{Sensitivity for $\beta$ on PACS.}
    \label{fig:short-beta-p}
  \end{subfigure}
    \hfill
  \caption{Sensitivity result for different values of $\alpha$ and $\beta$}
  \label{fig:ablation-alpha-beta}
\end{figure}

\begin{table}[t!]
    \centering
    \caption{Ablation study for $L^{GC}$ on Digits.}
\adjustbox{width=1\columnwidth}{
\begin{tabular}{cccc|ccccc}
\hline
$P$ & $P^G$ & $\mathcal{M}$ & $\mathcal{M}^G$ & SVHN & USPS & Synth & MNIST-M  \\ \hline
 \ding{56} & \ding{56}  & \ding{56} & \ding{56} & 25.51 & 70.54 & 35.01 & 54.61 \\
\ding{52} & \ding{56} & \ding{56} & \ding{56} & 38.30 & 68.92 & 52.75 & 83.94  \\
\ding{52} & \ding{52} & \ding{56} & \ding{56} & 44.55 & 76.39 & \textbf{60.94} & 85.23  \\
\ding{52} & \ding{52} & \ding{52} & \ding{56} & 43.28 & 72.29 & 57.52 & 83.21  \\
\ding{52} & \ding{52} & \ding{52} & \ding{52} & \textbf{47.19} & \textbf{78.66} & 59.89 & \textbf{85.74} \\ \hline
\label{tab:abla_lcp_d}
\end{tabular}}
\end{table}

\begin{table}[t!]
    \centering
    \caption{Ablation study for $L^{GC}$ on PACS.}
\adjustbox{width=0.85\columnwidth}{
\begin{tabular}{cccc|cccc}
\hline
$P$ & $P^G$ & $\mathcal{M}$ & $\mathcal{M}^G$ & Art & Cartoon & Sketch  \\ \hline
\ding{56} & \ding{56} & \ding{56} & \ding{56} & 44.75 & 16.94 & 26.16 \\
\ding{52} & \ding{56} & \ding{56} & \ding{56} & 62.21 & 26.57 & 36.64  \\
\ding{52} & \ding{52} & \ding{56} & \ding{56} & 63.67 & 28.71 & 38.32 \\
\ding{52} & \ding{52} & \ding{52} & \ding{56} & 64.10 & 30.29 & 40.66 \\
\ding{52} & \ding{52} & \ding{52} & \ding{52} & \textbf{65.01} & \textbf{30.91} & \textbf{41.19} & \\ \hline
\label{tab:abla_lgc_PACS}
\end{tabular}}
\end{table}

\noindent{\bf Sensitivity for $\alpha$ and $\beta$.} We show sensitivity analyses for the hyperparameters $\alpha$ and $\beta$ on two datasets, as depicted in Fig.~\ref{fig:ablation-alpha-beta}. The line plots include error bars representing standard deviations. Notably, the highest accuracies were achieved at values of 0.1 and 10 for $\alpha$, and 0.1 and 8 for $\beta$ on the Digits dataset and the PACS dataset, respectively.

In Table~\ref{tab:abla_lcp_d} and Table~\ref{tab:abla_lgc_PACS}, we show the ablation study for $\mathcal{L}^{GC}$ on Digits and PACS, considering the impact of four factors: $P$, $P^G$, $\mathcal{M}$, and $\mathcal{M}^G$.

Specifically, when simultaneously considering the four factors, the performance reaches the optimal level with the best classification accuracy on most targeted domains. Synth (60.94\%) performs better without class activation maps, mainly because the source domain is not informative enough to extract CAMs accurately. The PACS dataset is boosted in all settings when the informative photo is used as the source domain. This suggests that incorporating predictions and class activation maps, both local and global, has a positive impact on enhancing model generalization on the datasets.

\begin{figure}[t!]
  \centering
  \includegraphics[width=0.95\columnwidth]{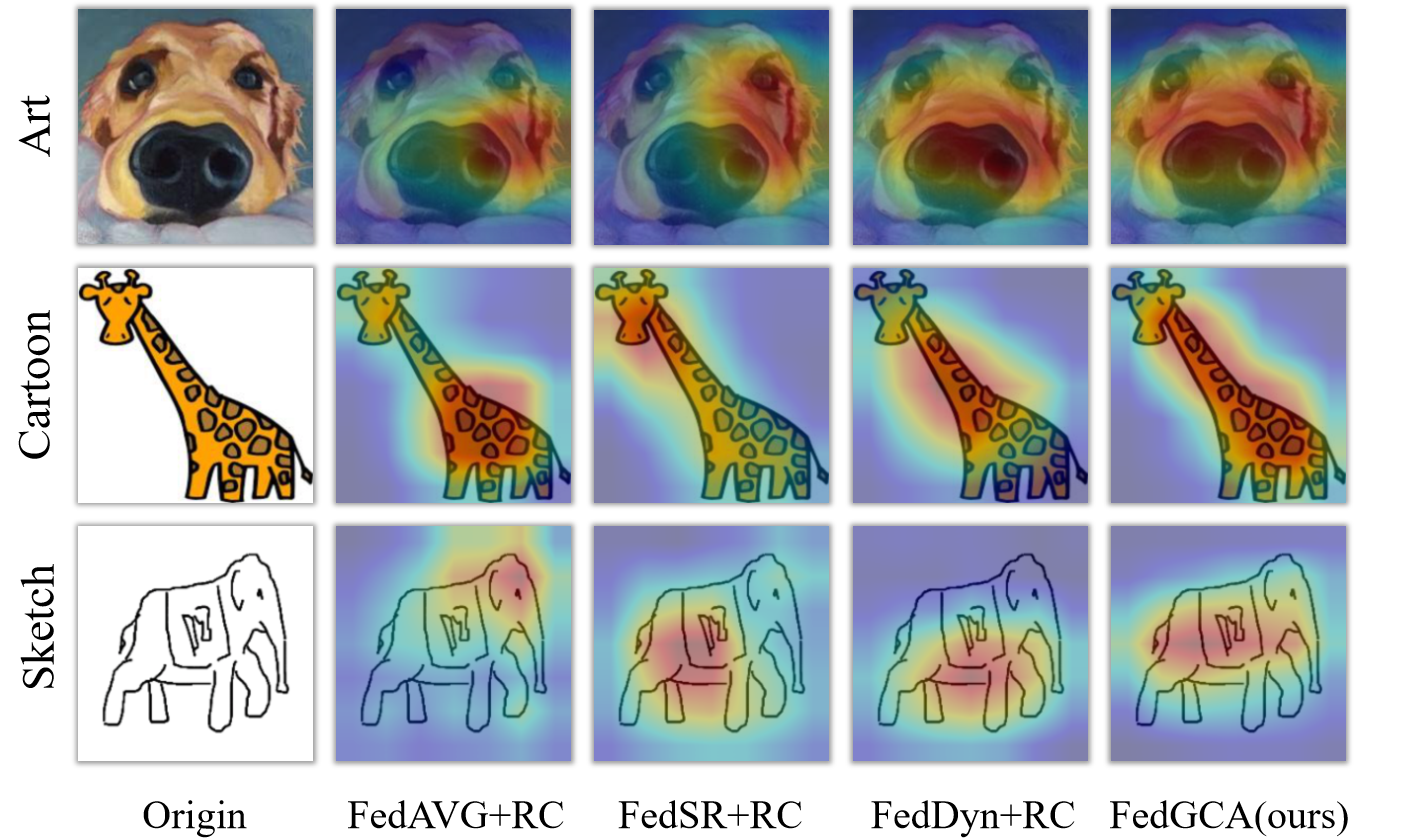}
  \caption{Visualization for CAMs of different methods.}
  \label{Fig:cam_show}
\end{figure}

\noindent{\bf Qualitative results.} In Fig.~\ref{Fig:cam_show}, we show CAMs for four different methods. It is observed that FedGCA can recognize and focus on the class-relevant objects in unseen domains. In detail, FedAvg cannot accurately identify category-critical regions. Adding regularizers (FedSR and FedDyn) improves the situation, but it is still not comprehensive. Our FedGCA makes it easier to locate the CAM.

\section{Conclusion}
In this paper, we have addressed the emerging challenges of the sFedDG problem, a practical scenario where clients are confined to a single domain, by designing the FedGCA method. Unlike FedDG settings, clients lack interactions with other domains, hindering the identification of common features crucial for effective generalization. To tackle the sFedDG challenge, we introduced a style-complement module that enriches the semantic information on each client. By employing the global guided semantic consistency strategy with dynamic regularization, we effectively addressed the limitations posed by the single-source and heterogeneous nature of sFedDG. Extensive experiments on different datasets successfully demonstrate the efficiency of FedGCA.


\bibliographystyle{IEEEtran}
\bibliography{main_icme2023}

\end{document}